\documentclass[sigconf]{acmart}
\AtBeginDocument{%
  \providecommand\BibTeX{{%
    \normalfont B\kern-0.5em{\scshape i\kern-0.25em b}\kern-0.8em\TeX}}}

\usepackage{latexsym}
\usepackage{graphicx}

\usepackage{amsmath}
\usepackage{cleveref}
\usepackage{xcolor}
\usepackage{enumitem}
\usepackage{tikz}
\usepackage{pgfplots}
\usepackage{subcaption}
\usepackage{url}
\usepackage{booktabs}
\usepackage{microtype}
\usepackage{placeins}
\usepackage{microtype}
\copyrightyear{2021}
\acmYear{2021}
\setcopyright{acmlicensed}\acmConference[SIGIR '21]{Proceedings of the 44th International ACM SIGIR Conference on Research and Development in Information Retrieval}{July 11--15, 2021}{Virtual Event, Canada}
\acmBooktitle{Proceedings of the 44th International ACM SIGIR Conference on Research and Development in Information Retrieval (SIGIR '21), July 11--15, 2021, Virtual Event, Canada}
\acmPrice{15.00}
\acmDOI{10.1145/3404835.3463099}
\acmISBN{978-1-4503-8037-9/21/07}
\begin{document}

\fancyhead{} 

\title{Cheap and Good? Simple and Effective Data Augmentation for Low Resource Machine Reading}

\author{Hoang Van}
\affiliation{%
  \institution{University of Arizona}
  \city{Tucson}
  \state{AZ}
  \country{USA}}
\email{vnhh@arizona.edu}

\author{Vikas Yadav}
\affiliation{%
  \institution{IBM Research}
  \city{Yorktown Heights}
  \state{NY}
  \country{USA}}
\email{vikasy@ibm.com}

\author{Mihai Surdeanu}
\affiliation{%
  \institution{University of Arizona}
  \city{Tucson}
  \state{AZ}
  \country{USA}}
\email{msurdeanu@arizona.edu}


\renewcommand{\shortauthors}{Van, Yadav, and Surdeanu}

\begin{abstract}
  We propose a simple and effective strategy for data augmentation for low-resource machine reading comprehension (MRC). Our approach first pretrains the answer extraction components of a MRC system on the augmented data that contains approximate context of the correct answers, before training it on the exact answer spans. The approximate context helps the QA method components in narrowing the location of the answers.  We demonstrate that our simple strategy substantially improves both document retrieval and answer extraction performance  by providing larger context of the answers and additional training data. In particular, our method significantly improves the performance of BERT based retriever (15.12\%), and answer extractor (4.33\% F1) on TechQA, a complex, low-resource MRC task. 
  Further, our data augmentation strategy yields significant improvements of up to 3.9\% exact match (EM) and 2.7\% F1 for answer extraction on PolicyQA, another practical but moderate sized QA dataset that also contains long answer spans.
\end{abstract}





\maketitle
\section{Introduction}
Supervised neural question answering (QA) methods have achieved state-of-the-art performance on several datasets and benchmarks \cite{seo2016bidirectional,devlin2018BERT,liu2019roBERTa,yadav2020unsupervised}. However, their success is fueled by large annotated datasets, which are expensive to generate and not always available in low resource settings. Moreover, machine reading comprehension (MRC) QA with span selection from a large set of candidate documents has been shown to require even more training data \cite{talmor2019multiqa,fisch2019mrqa,joshi2017triviaqa, dunn2017searchqa}. 

In this work, we propose a simple and effective data augmentation method that tackles the shortage of training data for MRC. The intuition behind our idea is straightforward: rather than directly train a neural approach on the answer spans provided during training, we start by {\em pretraining} it to identify the rough context in which the answer appears. Then we use this pretrained neural model in two different ways. First, instead of starting from scratch, we initialize the answer extraction model with the weights from this model before training it. Second, at inference time, we use the pretrained model as an additional document retrieval component: rather than extracting answers from any document, we focus only on documents that contain contexts identified as likely to contain an answer. 

\begin{figure}[t]
\includegraphics[width=0.8\linewidth]
{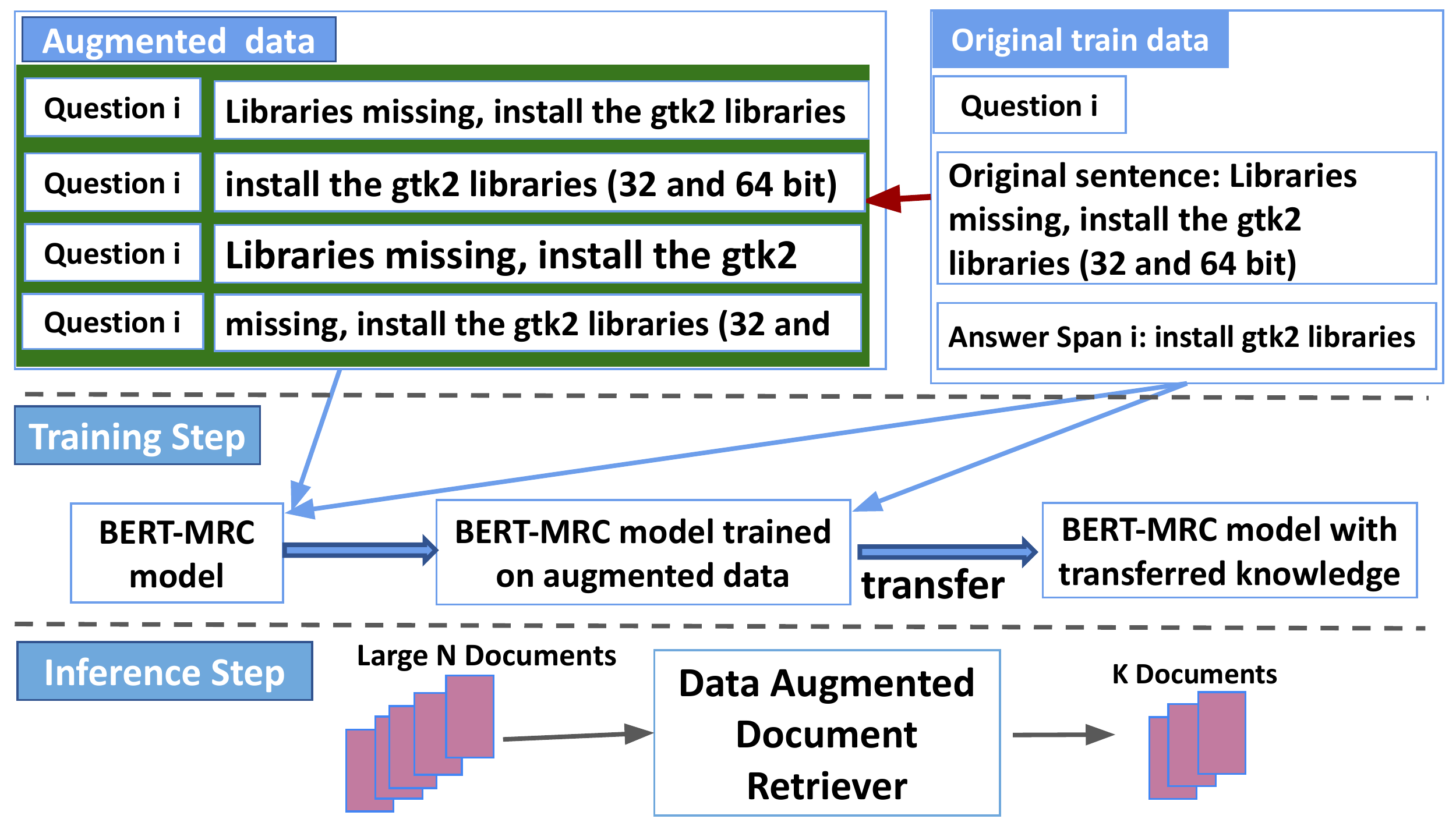}
\caption{\footnotesize Overview of our three key contributions: data augmentation (top), transfer learning (middle), and dynamic document retrieval (bottom). 
Top: we augment question-answer pairs with additional answer spans by moving the answer spans to left or right of the original spans by $d$ characters. 
Middle: we train the MRC answer extraction (AE) component with original settings on the combined augmented + original data. During transfer learning, the models trained on the combined data continue tuning on the original (without augmentation) data.
Bottom: at inference time, $k$ candidate documents are retrieved by the data-augmented document retriever (trained using the scheme above) from the pool of N documents per question. 
\label{fig:model}}
\vspace{-9mm}
\end{figure}

The key contributions of our work are:

{\flushleft {\bf (1)}} We introduce a simple yet effective data augmentation method for MRC QA. Our approach generates additional training data by artificially moving the boundaries of answer spans in the training partition, and training a model on this data to identify the approximate context in which the answer is likely to appear.

{\flushleft {\bf (2)}} We implement two different strategies to take advantage of the above model. (a) First, we transfer its knowledge into the answer extraction model by initializing with its learned weights, and (b) secondly, at inference time, we use the above model (trained on the fuzzy answer boundaries) to rerank documents based on their likelihood to contain an answer context, and employ answer extraction just in the top documents with the highest scores. 

{\flushleft {\bf (3)}} We demonstrate that utilizing approximate RC knowledge to both document retrieval and answer extraction results in considerable improvements in their performance on low resource RC tasks. In particular, after transferring the approximate RC knowledge, the BERT-based MRC model's performance is improved by 3.46\%, while the BERT-based document retriever's (\cite{castelli2019techqa}) performance is improved by 4.33\% (both in absolute F1 score) on TechQA, a real-world low-resource MRC dataset \cite{castelli2019techqa}. Further, our best model also outperforms the previous best document retrieval scores \cite{castelli2019techqa} by 15.12\% in document retrieval accuracy (DRA) 
on TechQA. Similarly, our proposed approach also yields 3.9\% EM and 2.7\% F1 improvements for the BERT-based answer extractor on PolicyQA, another practical and challenging QA dataset that has long answer spans  \cite{ahmad2020policyqa}. Our code and reproducibility checklist are publicly available.\footnote{{\url{https://github.com/vanh17/techqa} \\
$^\dagger$ Work performed while this author was a PhD student at the University of Arizona.}}



\section{Approach}
Fig.\ref{fig:model} highlights the three main components of our MRC approach: (a) data augmentation with fuzzy answer boundaries (top); (b)  transfer learning into the answer extraction component (middle); and (c) improved document retrieval. We detail these modules next.

\subsection{Data Augmentation} \label{sec:approach_augmentation}
We pose the data augmentation problem \cite{longpre2019exploration, liu2020tell} as a positive example expansion problem. 
First, we sample which answers should be used for augmentation with probability $p$. 
Second, once a question-answer pair is selected for augmentation, we generate $n$ additional fuzzy answer spans by moving the correct answer span to the left or right by $d$ characters. 
Each generated answer span becomes a new positive training data point. 
For example, for the answer {\em ``install gtk2 libraries''} in the sentence {\em``Libraries missing, install the gtk2 libraries (32 and 64 bit)''}, our data augmentation method generates the spans {\em``Libraries missing, install gtk2 libraries''} ($d = -19$) and {\em``install gtk2 libraries (32 and 64 bit)''} ($d = +16$) as additional answers (see Fig. \ref{fig:model}, data augmentation). 
Both $d$ and the number of augmented data points per training example are hyper parameters that we tune for the given tasks (see Section~\ref{sec:setup_augmentation}). Note that this augmented data is appended to the original training set.

Intuitively, a model trained on this artificial data (illustrated by the second BERT-MRC block in Figure \ref{fig:model}) would be capable of identifying the \textit{approximate} context in which an answer appears, even if it cannot locate the answer exactly \cite{ golub2017two}.\footnote{This is probably how students work towards answering reading comprehension questions as well. See, e.g.: {\tiny \url{https://tinyurl.com/khanrc}}.}
\vspace{-3mm}
\subsection{Transfer Learning} 
As shown in Figure \ref{fig:model}, we first train the BERT-MRC model on the augmented training data consisting of the fuzzy answer spans. Then, we transfer this knowledge of  approximate answer boundaries
by continuing to further train our BERT-MRC model on the original answer spans (without augmentation). The resulting model (the third BERT-MRC block in Fig. \ref{fig:model}) is the final model used for 
 answer span prediction in the complete task. 
 Note that, while we use BERT-MRC \cite{wang2019multi, wolf2019transformers} to train these answer extraction components, 
 same idea could be investigated with other answer extraction methods. 
 
\begin{table}[t]
\centering
\scalebox{0.6}{%
\begin{tabular}{l l c c c c}
\toprule
& Model & DRA@1 &  DRA@5 & F1 & Recall \\
\midrule
& \textbf{Baselines: Original Settings} & & & &\\
\midrule
1 & BERT Base (OBB) & 31.88 & 45.00 & 51.73 & 51.94 \\
2 & BERT Large (OBL) & 58.13 & 64.38 & 55.31 & 57.81\\
\midrule
& \textbf{Data Augmentation (DA)} & & & & \\
\midrule
3 & 60\% augmentation (aug\_60) & 46.25 & 51.25 & 53.52 & 53.67 \\
4 & 80\% augmentation (aug\_80) & \textbf{47.53*} & \textbf{53.00*} & \textbf{54.35*} & \textbf{54.73*} \\
5 & 100\% augmentation (aug\_100) & 46.33 & 48.75 & 53.56 & 52.76 \\
\midrule
& \textbf{DA+Transfer Learning} & & & & \\ 
\midrule
6 & aug\_60 + Transfer & 43.13 & 50.20 & 54.62 & 55.03\\
7 & aug\_80 + Transfer & \textbf{46.87*} & \textbf{52.10*} & \textbf{55.19*} & \textbf{55.68*} \\
8 & aug\_100 + Transfer & 42.58 & 49.75 & 54.11 & 53.39 \\
\midrule
& \textbf{Document Retrieval + OBL} & & & & \\
\midrule
9 & aug\_60 + OBL & 69.38 & 75.63 & 59.13 & 59.48 \\
10 & aug\_80 + OBL & \textbf{73.25*} & \textbf{78.13*} & \textbf{59.64*} & \textbf{59.84*} \\
11 & aug\_100 + OBL & 71.25 & 73.13 & 58.58 & 58.89 \\
\bottomrule
\end{tabular}}
\caption{\footnotesize Document retrieval accuracies (DRAs) and answer extraction results (F1, Recall) on the development partition of TechQA. We use BERT Base (similar to OBB) for all data augmentation and transfer learning experiments. OBL is a baseline in \cite{castelli2019techqa}.'Data Augmentation' shows results of models tuned with augmented data, starting with pretrained weights from BERT Base. 'Data Augmentation + Transfer Learning' lists results of data augmented models (DAM) after continuing to tune on the original (no augmentation) data. 'Document Retrieval + OBL' uses a hybrid approach where a model from the 'Data Augmentation' block is used to retrieve documents that are then fed to OBL for answer extraction. * indicates significant improvement over the respective baselines (p $<$ 0.02).}
\label{tab:document_retrieveal_accuracy_based_positive_examples}
\vspace{-8mm}
\end{table} 

\subsection{Document Retrieval}

Most MRC datasets include a list of $N$ candidate documents for each question \cite{kwiatkowski2019natural}. All these documents are scanned for possible answers by typical MRC approaches \cite{swayamdipta2018multi}. Instead, as shown in Fig.~\ref{fig:model},
we use only documents that are more likely to contain good answer contexts. In particular, we employ the model trained on the augmented data (with fuzzy answer boundaries) to produce the top $k$ answer spans with the highest scores, and keep only the documents associated with these answer spans. 
For example, on TechQA, by selecting the top documents containing the high score answer spans, our approach narrows down the number of candidate documents to an average of $6.41$ per question\footnote{We also observed that the maximum number of candidate documents is $10$} from the initial pool of 50 documents. As shown in Tab.\ref{tab:performance_policy} 
, feeding a smaller but more relevant pool of candidate documents leads to substantial improvements in the AE performance.

\begin{table}[t]
\centering
\scalebox{0.70}{
\begin{tabular}{l l c c}
\toprule
& Models & F1 &  EM \\
\midrule
& \textbf{Baselines} & \\
\midrule
1 & OBB pretrained on SQuAD (OBS) & $57.2^{\dag}$ & $27.5^{\dag}$ \\
\midrule
& \textbf{Data Augmentation} & &\\
\midrule
2 & OBS + 3\% augmentation (OBS\_aug\_3) & 58.8 & 29.3 \\
3 & OBS + 4\% augmentation (OBS\_aug\_4) & \textbf{59.8*} & \textbf{30.8*} \\
4 & OBS + 5\% augmentation (OBS\_aug\_5) & 59.2 & 30.2 \\
\midrule
& \textbf{Data Augmentation + Transfer Learning} & & \\
\midrule
5 & OBS\_aug\_3 + Transfer & 59.2 & 29.9 \\
6 & OBS\_aug\_4 + Transfer & \textbf{59.9*} & \textbf{31.4*} \\
7 & OBS\_aug\_5 + Transfer & 59.3 & 30.6 \\
\bottomrule
\end{tabular}}
\caption{\footnotesize Answer extraction results on PolicyQA development. * indicates significant improvement over the baseline (line 1) with p $<$ 0.03. \dag ~indicates experiments from a different paper that we reproduced for a fair comparison. These results differ from the original paper \cite{ahmad2020policyqa}, which is likely due to the different hardware and library versions used.}
\label{tab:performance_policy}
\vspace{-10mm}
\end{table}

\section{Experimental Setup} 
We evaluate our approach on two complex MRC tasks: TechQA \cite{castelli2019techqa}\footnote{\url{https://github.com/IBM/techqa}} and PolicyQA \cite{ahmad2020policyqa}\footnote{\url{https://github.com/wasiahmad/PolicyQA}}. Both datasets contain questions from {\em low-resource domains}, and their questions are answered by {\em long text spans}, which makes these tasks more challenging for RC approaches \cite{castelli2019techqa}. Specifically, TechQA is a real-world dataset that requires extraction of long answer spans to industrial technical questions. TechQA questions and answers are substantially longer than those found in common datasets \cite{rajpurkar2018know}. The average answer span and average question length in the training partition of this dataset are 48.1 and 52.1 tokens, respectively. 
Further, this task is relatively low resourced: the training partition contains only 600 questions; the development one contains 310. Each question is associated with 50 documents that are likely to contain the answer. Note that some questions (25\% in training, 48\% in development) are unanswerable i.e., they do not have a answer in the 50 documents provided.
Similar to TechQA, PolicyQA is a another real-world MRC task that requires the extraction of answers to privacy policy questions from long documents (average document length is 106 tokens). However, each PolicyQA question is associated with just 1 given document. But, similar to TechQA, the average answers spans in PolicyQA are also long: the average answer span and average question length in the training partition of this dataset are 11.2 and 13.3 tokens, respectively. Please note that we selected PolicyQA as the second dataset to demonstrate that our approach works with moderate resourced datasets\footnote{PolicyQA has 17K questions in the training partition.} as well.

For reproducibility, we use the default settings and hyper parameters recommended by the task creators~\cite{castelli2019techqa, ahmad2020policyqa}. Through this, we aim to separate potential improvements of our approaches from those made with improved configurations. 
Due to computational constraints, we report transfer learning results using BERT Base (uncased) from the HuggingFace library\footnote{\url{https://github.com/huggingface}}.
Additionally, for TechQA, we combine our document retrieval strategy with the answer extraction component based on BERT Large (uncased whole word masking tuned on SQUAD \cite{rajpurkar2018know}) that is provided by the organizers.

\subsection{Evaluation Measures} \label{sec:metrics}
\paragraph{Answer extraction:} we directly followed the evaluation metrics proposed by the original task creators \cite{castelli2019techqa, ahmad2020policyqa}. Specifically, we used the following ancillary metrics to evaluate the answer extraction components: F1 (TechQA, PolicyQA) and exact match (EM) (PolicyQA). 
For each question, the F1 score is computed based on the character overlap between the predicted answer span and the correct answer span.
 We also report the Recall score on TechQA, similarly to the task organizers \cite{castelli2019techqa}.
\vspace{-1mm}
\paragraph{Document retrieval:} we report the original evaluation metric, i.e., document retrieval accuracy (DRA), as proposed by the task organizers \cite{castelli2019techqa}\footnote{We do not use Precision@$N$ score for retrieval evaluation to keep our results comparable with the other original methods evaluated on these datasets.}.
 In MRC tasks such as TechQA, DRA is calculated only over $answerable$ questions. The score is 1 if the retrieval component retrieves the document containing the correct answer span, and 0 otherwise. We report the performance with DRA@1 and DRA@5, which score the top 1 and 5 retrieved documents, respectively.

\subsection{Hyper parameter tuning}
\label{sec:setup_augmentation}
We tuned three hyper parameters for data augmentation: percentage of augmented training data points, span displacement size, and number of new spans generated per correct answer span. On TechQA, we tried augmenting 20, 40, 60, 80, 100 \% of the questions from training data and found 80\% of the training data as the best threshold. For each (randomly) selected question from the training data, we generated 6 augmented data points by moving the gold answer span boundaries. Specifically, we moved answer spans up to 15 characters to left and up to 20 characters to the right for TechQA. On PolicyQA, we tuned smaller movements of answer spans\footnote{because PolicyQA answer spans are smaller than those in TechQA.} such as 5 characters on the left and 10 characters on the right of the gold answer spans. For each PolicyQA question, similar to TechQA, we generated 6 additional data points based on the scheme above.
We tuned these three hyper parameters using DRA@5 (TechQA) and F1 (PolicyQA),\footnote{PolicyQA does not have document retrieval components because each question is associated with only one long document.} with the aims of improving the ability to localize the context in which an answer appears, and reducing the likelihood of missing correct documents.\footnote{As expected, tuning these hyper parameters is a balancing act between the desire to add more training data and introducing noise. We found that there is a "Goldilocks" zone for key parameters such as $d$, and the percentage of augmented training data.}


\begin{table*}[t]
\scalebox{0.75}{%
\begin{tabular}{l l l l l}
\toprule
& \textbf{Question} & \textbf{Context} \\
\midrule
1 & Will you take my consent before  & E-mail and newsletters: \textbf{*If you signed up*} for one of our email collecting or using my data? newsletters, we will send you \\
 & & ... We may also use the information ... (\textbf{\#}if you have opted in for this information.\textbf{\#})  \\
\midrule
2 & Does the company use any form of & We reserve the right to use, transfer, and \textbf{*}share \textbf{\#}\textbf{aggregated\#, anonymous} \textbf{data*} about our users as a group for any lawful \\
 & aggregated or anonymized information?  & business purpose, such as analyzing usage trends and seeking compatible advertisers, sponsors, clients and customers.\\
\bottomrule
\end{tabular}}
\caption{\footnotesize Qualitative comparison of the outputs from our approach (Data Augmentation + Transfer Learning) and the respective baseline (Original settings with BERT base pretrained on SQUAD) on PolicyQA. The text in bold font indicates the gold answer for the corresponding question. The text between \textbf{\#} and \textbf{*} indicate the answers from the baseline and our approach respectively for the given question.}
\label{tab:example_policy}
\vspace{-7mm}
\end{table*}

\begin{table*}[t]
\scalebox{.7}{%
\begin{tabular}{l l l l}
\toprule
 & \textbf{Gold Data} & \textbf{Our Approach} & \textbf{Baseline} \\
\midrule
 1 & \textbf{Q:} What is the difference between Advanced and & \textbf{Doc:} None, \textbf{Ans:} None & \textbf{Doc:} swg21672885 \\
 
    &  AdvancedOnly options \textbf{Doc:} None, \textbf{Ans:} & & \textbf{Ans:} The bootstrapProcessServerData ... \\
\midrule
 2 & \textbf{Q:} Is using a monitored JBoss a Windows server & \textbf{Doc:} swg21967756 & \textbf{Doc:} swg21967756 \\
 
   &  with ITCAM supported in  Service? & \textbf{Ans:} The JBoss service is not available & \textbf{Ans:} The JBoss service is not available \\
   
   & \textbf{Doc:} swg21967756 & to run as a Windows service when configured & \\
   
   & \textbf{Ans:} The JBoss service is not available to run with the & with the ITCAM for J2EE agent/DC & \\
   
   & ITCAM ... involves changes ... files currently not supported. & because this involves changes & \\
\midrule
3 & \textbf{Q:} Why do we get server error message when running & \textbf{Doc:} swg21982354 & \textbf{Doc:} swg21981881\\

  & BIRT reports after upgrading to Atlas 6.0.3.3? & \textbf{Ans:} This happens when the BIRT Reports is running in  & \textbf{Ans:} In Atlas 6.0.3.3, the Atlas Extensions logging \\
  
  & \textbf{Doc:} swg21982354 & Standalone mode and happens due to a new configuration  & configuration has moved to log4j.properties file.\\
  
  & \textbf{Ans:} This happens when the BIRT Reports is & - report.standalone.userid 1. Navigate to Atlas Properties & 1. Navigate to Atlas\_Install\_folder/Atlas/Properties \\
  
  & running in Standalone ... report.standalone.userid & Folder 2. Edit ... \#report.standalone.userid=1 & 2. Edit log4.properties file 3. Update the path  \\
  
\bottomrule
\end{tabular}}
\caption{\footnotesize Qualitative comparison of the outputs from our approach (Data Augmentation + Transfer Learning) and the respective baseline (OBB) on TechQA.}
\label{tab:example_tech}
\vspace{-9mm}
\end{table*}

\section{Results and Discussion}
We investigate the contribution of our ideas on document retrieval and AE performance as explained below.

\subsection{Document Retrieval}

Table \ref{tab:document_retrieveal_accuracy_based_positive_examples} shows that data augmentation helps with the document retrieval task in TechQA. Our approaches improve DRA scores considerably compared to the BERT Base baseline (OBB) that serves as the starting point for our experiments (row 1 in Table \ref{tab:document_retrieveal_accuracy_based_positive_examples}).
For example, the 'Data Augmentation' models, which were trained on the data augmented with fuzzy answer boundaries (rows 3--5 in Table \ref{tab:document_retrieveal_accuracy_based_positive_examples}), improve DRA@1 by 14.37--15.65\% (absolute) compared to OBB.
Similarly, the 'Data Augmentation + Transfer Learning' models (rows 6--8 in Table \ref{tab:document_retrieveal_accuracy_based_positive_examples}), which continued to train on the original data, improve DRA@1 by  10.70--14.99\% (absolute).

Interestingly, 'Data Augmentation' models perform better for document retrieval than their equivalent 'Data Augmentation + Transfer Learning' models. 
We hypothesize that the transfer learning phase encourages the models to focus on the answer extraction task, ignoring the larger textual context infused in the previous step, which is useful for document retrieval. To further analyze the effects of data augmentation on document retrieval, we also list the results of a hybrid approach, where the documents retrieved by our 'Data Augmentation' models are used as input to the OBL baseline from \cite{castelli2019techqa}. DRA@1 increases noticeably by 11.25--15.12\% (subtract line 2 from lines 9--11 in Table \ref{tab:document_retrieveal_accuracy_based_positive_examples}). We conclude that the candidate documents retrieved by data augmented models can boost the document retrieval performance of an external, independent model.

To illustrate the benefits of additional augmented data, we provide a qualitative comparison between the outputs of our approach and the respective baseline in Table \ref{tab:example_tech}. With the transferred knowledge of larger context, our model successfully retrieves the correct document (row 3 in Table \ref{tab:example_tech}). Further, our method correctly detects if a question has no answer in the given documents (row 1 in Table \ref{tab:example_tech}). The baseline struggles in both scenarios.

\subsection{Answer Extraction}
\paragraph{Data augmented models:} 
As shown in Table \ref{tab:document_retrieveal_accuracy_based_positive_examples}, our data-augmented models (DAMs) yield improved AE performance on TechQA. Compared to the OBB baseline (row 1 in Table \ref{tab:document_retrieveal_accuracy_based_positive_examples}), F1 scores for answer extraction increase when data augmentation is added (see rows 3--5 in Table \ref{tab:document_retrieveal_accuracy_based_positive_examples}).  There are absolute increases of 1.79--2.62\% in F1 for DAMs. Table \ref{tab:example_tech} (row 2) shows a qualitative example where the additional knowledge from the larger context helps the answer extraction component provide better (but not perfect) answer boundaries. Further, there are absolute increases of 3.82\%, 4.33\%, and 3.27\% in F1 scores, respectively, when the DAMs are used to retrieve candidate documents as inputs to OBL for TechQA (rows 9--11 in Table \ref{tab:document_retrieveal_accuracy_based_positive_examples}). 
These results confirm that the better documents retrieved by our method lead to better answers extracted by OBL. Importantly, even though our document retrieval component produces much fewer documents than the original of 50, answer recall increases across the board, further confirming the improved quality of the documents retrieved.

Similarly, on PolicyQA, data augmentation  improves AE performance (see Table \ref{tab:performance_policy}). Compared to the OBS baseline (row 1), evaluation scores for answer extraction increase when data augmentation is added (absolute increases of 1.6--2.6\% and 1.8--3.3\% in F1 and EM, respectively, see rows 2--4)\footnote{Please note that since PolicyQA contains 17,056 questions in its training partition (which is 30  times more than the TechQA training set),  we observed the best improvements on PolicyQA when only 4\% of the training data were used to generate new QA pairs with approximate boundaries.}. 
This suggests that our simple approach is not dataset specific, which is encouraging. Similar to other previous works, we conjecture that our proposed data augmentation approach contributes substantially in very low resource tasks (e.g., TechQA), and its impact decreases as more training resources are available (e.g., PolicyQA). 

\textit{Transfer learning models:} Compared to OBB, our 'Data Augmentation + Transfer Learning' (\texttt{DA+TL}) models perform considerably better for answer extraction in TechQA (rows 6--8 in Table \ref{tab:document_retrieveal_accuracy_based_positive_examples}). Transfer learning models also outperform data augmented models (0.55--1.1\% absolute increases in F1). Comparing the best configurations between the two (rows 4 vs. 7 in Table \ref{tab:document_retrieveal_accuracy_based_positive_examples}), we observe that transfer learning models extract answer spans more effectively (absolute increases of 0.84\% in F1 score). This confirms that transferring to the exact answer spans is beneficial for answer extraction components.

On PolicyQA, our \texttt{DA+TL} models improve AE performance (2.0--2.7\% and 2.4--3.9\% absolute increases in F1 and EM, subtract row 1 from rows 5--7 in Table \ref{tab:performance_policy}). Also, Transfer Learning further improves AE performance (absolute increases of 0.1\% and 0.6\% in F1 and EM, compare the best configurations between the two: rows 4 vs. 7 in Table \ref{tab:performance_policy}). Table \ref{tab:example_policy} shows qualitative examples comparing \texttt{DA+TL} with the OBS baseline. Our qualitative analysis shows that our \texttt{DA+TL} approach more efficiently: (a) locates the correct answers spans (see row 1 in Table \ref{tab:example_policy}), and (b) captures necessary information by expanding the answer span (row 2 in Table \ref{tab:example_policy}).

\section{Conclusion} 
We introduced a simple strategy for data augmentation for MRC, where we generate artificial training data points where answer spans are moved to the left/right of the correct spans. We demonstrated that this simple idea is useful for document retrieval (because it was able to capture the larger context in which the answer appears), and also as an effective pretraining step for answer extraction (because of the additional training data, which helps the extraction model narrow its search space). Our combined contributions yield the best overall performances for TechQA and PolicyQA.

\section{Acknowledgements}
\footnotesize{
This work was supported by the Defense Advanced Research Projects Agency (DARPA) under the World Modelers program, grant number W911NF1810014. Mihai Surdeanu declares a financial interest in lum.ai. This interest has been properly disclosed to the University of Arizona Institutional Review Committee and is managed in accordance with its conflict of interest policies.
}

\bibliographystyle{ACM-Reference-Format}
\bibliography{sample-base}
\end{document}